\documentclass[conf]{new-aiaa}
\usepackage[utf8]{inputenc}

 \usepackage[version=4]{mhchem}

\usepackage{ifpdf}
\usepackage[pdftex]{graphicx}
\usepackage{amsmath}
\usepackage{amssymb}
\usepackage{algorithm}
\usepackage[noend]{algpseudocode}
\usepackage{array}
\usepackage{subcaption}
\usepackage{url}
\usepackage{multicol}
\usepackage{booktabs}
\usepackage{siunitx}
\usepackage{color}

\title{Heterogeneous Fixed-wing Aerial Vehicles for Resilient Coverage of an Area}

\author{Sachin Shriwastav\footnote{Graduate Student, Mechanical Engineering, sachins@hawaii.edu} and Zhuoyuan Song\footnote{Assistant Professor, Mechanical Engineering, zsong@hawaii.edu}}
\affil{University of Hawai'i at M\=anoa, Honolulu, HI 96822}

\begin{document}

\maketitle

\begin{abstract}
This paper presents a distributed approach to provide persistent coverage of an arbitrarily shaped area using heterogeneous coverage of fixed-wing unmanned aerial vehicles (UAVs), and to recover from simultaneous failures of multiple UAVs. The proposed approach discusses level-homogeneous deployment and maintenance of a homogeneous fleet of fixed-wing UAVs given the boundary information and the minimum loitering radius. The UAVs are deployed at different altitude levels to provide heterogeneous coverage and sensing. We use an efficient square packing method to deploy the UAVs, given the minimum loiter radius and the area boundary. The UAVs loiter over the circles inscribed over these packing squares in a synchronized motion to fulfill the full coverage objective. An top-down hierarchy of the square packing, where each outer square (super-square) is partitioned into four equal-sized inner squares (sub-square), is exploited to introduce resilience in the deployed UAV-network. For a failed sub-square UAV, a replacement neighbor is chosen considering the effective coverage and deployed to the corresponding super-square at a higher altitude to recover full coverage, trading-off with the quality of coverage of the sub-area. This is a distributed approach as all the decision making is done within close range of the loss region, and it can be scaled and adapted to various large scale area and UAV configurations. Simulation results have been presented to illustrate and verify the applicability of the approach.

\end{abstract}

\section{INTRODUCTION}
Unmanned aerial vehicles (UAVs) are effective and efficient tools for coverage and sensing applications. Among various types of vehicles, fixed-wing UAVs are mostly used for mapping and surveillance by sweeping through an area for data collection and then returning to a base for the processing of the data. On the other hand, the rotor-type (e.g., quadcopters or hexacopters) can be used for persistent coverage and sensing scenarios. However, the biggest limitation of rotor systems is their endurance, as they have constraints on their payload and hence the power source they can carry. Compared to rotor vehicles, fixed-wing UAVs consume less energy to achieve and sustain mobility and are able to maintain airborne for longer duration~\cite{oettershagen2016long, guo2011development, oettershagen2016perpetual}. Their oftentimes large wing surface areas allow installation of solar panels that may further extend their flight time. Fixed-wing UAVs are suitable candidates for sensing and coverage. However, fixed-wing UAVs' mobility is typically limited by their minimum cruise speeds and loitering circles, making the coordination of UAV fleet challenging~\cite{Song:17a, Song:18d}.

Multi-UAV coverage is a well-researched area and boasts various methods for using fixed-wing UAVs for various applications that involve coverage control or optimization. Fixed-wing UAVs are traditionally used for fly-by coverage and the problem addressed can mostly be categorised as efficient path planning. These approaches take into account the characteristics, dynamics, and properties of the vehicles and the environment to improve efficiency. Chen et al.~\cite{chen2014coverage} reviewed the existing coverage algorithms in UAV network, and classified those algorithms into different groups according to factors considered, like coverage ability, mobility, lifetime, connectivity, and obstacles. Schleich et al.~\cite{schleich2013uav} proposed a decentralized and localized algorithm to control the mobility of multiple UAVs, i.e. a fleet, offers various advantages compared to the single UAV scenario, such as longer mission duration, bigger mission area or the load balancing of the mission payload. The network connectivity was maintained via a tree-based overlay network, of which root was the base station of the mission, and was created by predicting the future positions of one-hop neighbours. Mozaffari et al.~\cite{mozaffari2016efficient} used an efficient deployment of multiple UAVs acting as wireless base stations that provide coverage was analyzed by the ground users. Following this, the downlink coverage probability for UAVs was derived as a function of the altitude and the antenna gain. Next, using circle packing theory, the 3-D locations of the UAVs were determined in a way that the total coverage area was maximized while maximizing the coverage lifetime of the UAVs. Avellar et al.~\cite{avellar2015multi} presented an algorithm for minimum-time coverage of ground areas using a group of UAVs equipped with image sensors by modeling the area as a graph and solving a mixed-integer linear programming problem. Wei et al.~\cite{wei2017scaling} addressed the capacity and delay of the UAV network with $n$ UAVs that are intended to monitor three-dimensional environment such as air pollution, toxic gas leakage, etc. They derived the capacity and delay scaling laws of UAV network with mobility and pattern information. Xu et al.~\cite{xu2011optimal} presented an adaptation of an optimal terrain coverage algorithm for an aerial application. The general strategy involves computing a trajectory through a known environment with obstacles and ensures complete coverage of the terrain while minimizing path repetition. The paper introduced a system that applies and extends this generic algorithm to achieve automated terrain coverage using an aerial vehicle. Paull et. al~\cite{paull2013sensor} presented an algorithm where area coverage with an on-board sensor was an important task for a UAV while maintaining an in-situ coverage map based on its actual pose trajectory and making control decisions based on that map. Chen et al.~\cite{chen2018self} presented a self-organized, distributed and autonomous approach for sensing coverage for multiple UAVs with an approach that takes into account the reciprocity between neighboring UAVs to reduce the oscillation of their trajectories. Nedjati et al.~\cite{nedjati2016complete} presented a post-earthquake response system for rapid damage assessment. In this system, multiple UAVs were deployed to collect images from an earthquake site and create a response map for extracting useful information. Coombes et al.~\cite{coombes2017boustrophedon} addressed the need for an enhanced understanding of the wind effects on fixed-wing aerial surveying, and used Boustrophedon paths based on sweep angle relative to the wind that minimises the flight time. In \cite{coombes2018fixed}, the algorithm took into account environmental factors and aircraft dynamics. By decomposing the complex survey regions into many smaller arrangements, Boustrophedon paths can be used to cover them. Darbari et al.~\cite{darbari2017dynamic} presented a dynamic path planning algorithm for a UAV surveying a cluttered urban landscape. Voronoi Tessellation of the search space and identification of key waypoints in the form of milestones lead to an efficient mapping of the region to be surveyed. The changes in the environment were handled effectively by the decision process in the form of a local or global planner. The application of 3D Dubin's curve leads to smooth and dynamically feasible trajectories at runtime. In \cite{varga2015distributed}, teams of fixed-wing micro-aerial vehicles (MAVs) could provide a wide area coverage and relay data in the wireless ad-hoc networks. The authors proposed a distributed control strategy that is based on the attraction and repulsion between MAVs and relies only on local information. Ahmadzadeh et al.~\cite{ahmadzadeh2006multi} presented an algorithm for time-critical cooperative surveillance with a set of unmanned aerial platforms using an Integer Programming (IP)-based strategy for feasible trajectories while incorporating the complexity and coupling of the camera fields of view and flight paths. In \cite{ahmadzadeh2006cooperative}, the authors addressed the generation of team flight plans and controllers that enable a heterogeneous team of UxVs (x: A-Aerial, G-Ground) to maximize the spatio-temporal coverage while satisfying hard constraints such as collision avoidance and positional accuracy. Danoy et al.~\cite{danoy2015connectivity} presented an online and distributed approach for bi-level flying ad-hoc networks, in which the higher-level fixed-wing fleet serves mainly as a communication bridge for the lower-level fleets that conduct precise information sensing. In \cite{sachin2020coordinated}, full coverage of a desired rectangular area was maintained using homogeneous fixed-wing aerial vehicles loitering on a given altitude, using both square and hexagon packing. In the case of multiple UAV failures, the algorithm restored full coverage by modifying the loitering locations and a new homogeneous radius of the remaining UAVs at the same altitude. Savla et. al~\cite{savla2007coverage} studied a facility location problem for groups of Dubin’s vehicles, non-holonomic vehicles constrained to move along planar paths of bounded curvature, without reversing direction. Given a compact region and a group of Dubin's vehicles, the coverage problem is to minimize the worst-case traveling distance.

This paper presents an approach to use complete, and potentially persistent coverage of an arbitrary area using the minimum number of fixed-wing UAVs, while discussing the resilience aspect of the deployed UAV network over agent failures. The proposed algorithm defines a multi-level square packing over the given area of arbitrary shape, and uses a patrolling fleet of fixed-wing UAVs in pre-specified loiter circles over the squares for obtaining full coverage. The UAVs at a given altitude level have homogeneous coverage and are synchronized in their motion while on the same level; it is thus called \textit{level-homogeneous deployment}. The algorithm also incorporates resilience, that is, when one or more sub-square neighbors of a UAV fail, it is eligible to transit to the super square and loiter at the respective circle at the higher altitude. The choice of the UAV to transit to the super-square is based on its effective coverage and the transition is made using Dubin's path algorithm~\cite{savla2007coverage, lugo2014dubins, mclain2014implementing}. The proposed algorithm relies on the reference literature for obtaining the near-optimal length of the Dubin's path, and uses the well-researched method to obtain synchronized transit and thus synchronized motion at the higher level. It is a distributed and scalable approach as the detection of failure and the handling for coverage recovery are locally occurred and managed processes. As all the communication and decision making are limited within the immediate super-square, the proposed approach reduces the communication load of the network. Various arbitrary areas and failure scenarios were tested to verify the applicability and efficacy of the proposed approach in terms of efficient coverage and fault tolerance. This method is also scalable in terms of area size, and adaptive for various boundaries as well as the physical and mechanical limitations of the vehicles.

 We introduce coverage resilience in the coordination algorithm, which defines that in case of failure of their super-square neighbors, a UAV breaks off from loitering at the sub-square level and traces a short path using Dubin's path algorithm~\cite{savla2007coverage, lugo2014dubins, mclain2014implementing} to move to and loiter at the parent-square level. The proposed algorithm relies on the reference literature for obtaining the near-optimal length of the Dubin's path, as this is not the focus of this paper. It is a distributed and scalable approach as the detection of failure and the handling for coverage recovery are locally occurred and managed processes. As all the communication and decision making are limited within the immediate super-square, the proposed approach reduces the communication load of the network. The proposed method also focuses on covering an arbitrary geometric using level-homogeneous squares of different sizes at different altitude levels. This, in itself, is an interesting outcome as the arbitrary area is fully covered while minimizing the overlap between the coverage footprints of the neighboring UAVs. Various arbitrary areas and failure scenarios were tested to verify the applicability and efficacy of the proposed approach in terms of efficient coverage and fault tolerance. This method is also scalable in terms of area size, and adaptive for various boundaries as well as the physical and mechanical limitations of the vehicles. 

The major contributions of this work are as follows:
\begin{enumerate}
    \item The proposed algorithm ensures full coverage of the desired area using a fleet of fixed-wing UAVs, flying at different altitudes;
    \item The proposed method also focuses on covering an area of arbitrary geometric shape using level-homogeneous squares of different sizes at different altitude levels. It can also handle no-deployment zones within the area. This, in itself, is an interesting outcome as the arbitrary area is fully covered while minimizing the overlap between the coverage footprints of the neighboring UAVs;
    \item The presented algorithm provides coverage resilience by addressing coverage recovery problem in case of simultaneous multiple UAV failures, to still maintain full coverage with reduced resources;
    \item It is a distributed and scalable approach. As the recovery decision is local, the maintenance effort is reduced.
\end{enumerate}

The remainder of the paper is organized as follows: Section~\ref{section:prelim} presents the preliminaries and assumptions, Section~\ref{section:approach} discusses the details of the proposed approach, Section~\ref{section:sims} presents the simulation results and discussions. Finally, Section~\ref{section:last} lists some of the possible future work in this domain and concludes the paper.

\section{PRELIMINARIES}
\label{section:prelim}
The area to be covered by the UAV fleet can be represented as a graph with the vertices representing the longitude and latitude. The deployed UAVs would represent a subset of the graph nodes $\mathbb{V}'$ as a sub-graph $\mathbb{G}' = (\mathbb{V}',\mathbb{E}')$, and the edges between the neighboring UAVs ($\mathbb{E}'$) represent the active communication link.

\subsection{Assumptions}
The following assumptions are made to simplify the analysis without the loss of generality:
\begin{enumerate}
\item The UAVs are homogeneous, that is, they have the same size, weight, minimum turn radius, and communication and sensing capabilities;
\item The cruising velocity is constant and uniform for all the UAVs;
\item Each UAV knows its location at any point in time;
\item The UAVs are automatically able to communicate with any other UAV within its communication range ($r_\text{com}$), and the communication is isotropic, that is, they can communicate across the altitude levels.
\end{enumerate}

\subsection{Definitions}
\begin{figure}
    \centering
    \includegraphics[width=0.7\linewidth]{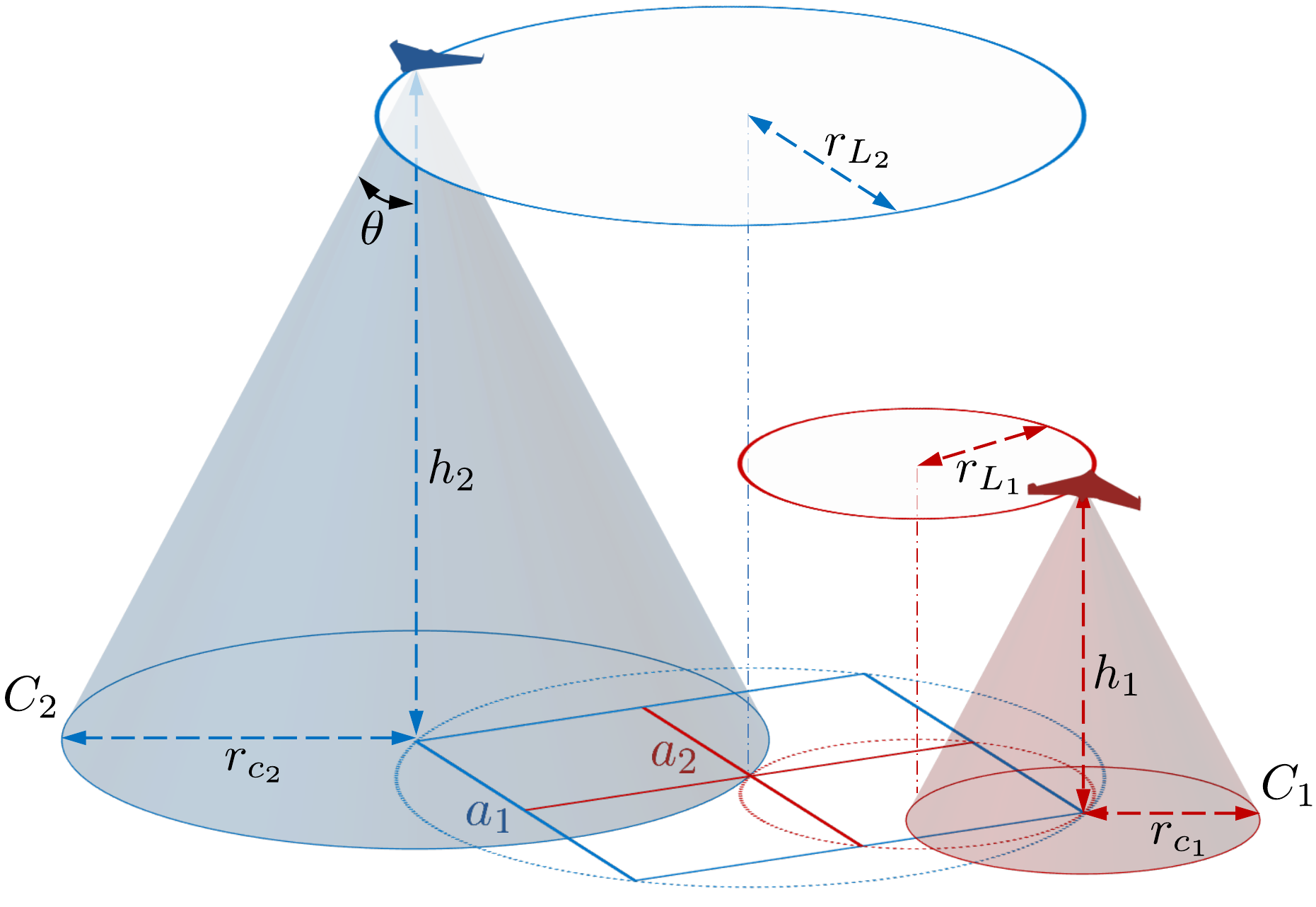}
    \caption{Parameters for the proposed approach: Level-homogeneous UAV deployment. The loiter circles and their instantaneous coverage footprint for two UAVs at different altitudes are shown.}
    \label{fig:illustration}
\end{figure}

Following are the key introductory concepts for the presentation of the proposed method~\cite{sachin2020coordinated}. These quantities have been graphically summarized in Fig.~\ref{fig:illustration}.

\noindent \textbf{Altitude level ($h_i$):} The altitude level represents the height from all the permitted height values in the square packing, where the subscript $i$ represents the instantaneous altitude level, and $h_i \in \{ h_1, \ldots , h_n\}$, for the $n$ pre-specified altitude levels. In this paper, we have limited the levels to $n=4$ to maintain acceptable quality of coverage. In Fig.~\ref{fig:illustration}, $h_1$ and $h_2$ represent two sequential altitude levels. As all the UAVs show similar behavior (e.g., loiter radius, velocity, etc.) on a given level, this approach is \textit{level-homogeneous}.

\noindent \textbf{Field of view (FOV):} The FOV is a physical property of the sensor being used by the UAV and defines the coverage footprint based on the altitude of the platform. In Fig.~\ref{fig:illustration}, The FOV has been marked by $\theta$. Based on the sense used, the sensing quality ($q$) can be defined as $q_i \propto \ {1}/{h_i}$, which means that the coverage quality decreases linearly as the altitude increases, and vice-versa.

\noindent \textbf{Phase synchronization:} For a UAV loitering at a given altitude level, the phase has been defined in this paper as the angle at which they are. It has been shown in Fig.~\ref{fig:illustration} as $\phi$. We assume that the motion is synchronized, that is, all the loitering UAVs have the same phase at every instant of time for maximum separation, and hence the largest effective coverage.

\noindent \textbf{Loiter radius ($r_\text{l}$):} Any form of a fixed-wing vehicle has constraints on maneuverability and it cannot stay stationary while it is airborne. Instead, it can fly in a circle over the region of interest, called the \textit{loiter circle}. The radius of that circle at a given instant is called the loiter radius. In Fig.~\ref{fig:illustration}, the loiter radius for two different altitude levels have been shown by the radii $r_{\text{L}_1}$ and $r_{\text{L}_2}$.

The physical properties and cruising velocity of a fixed-wing UAV system define the lowest value of the loiter radius, called the minimum turning radius~\cite{mclain2014implementing}, given by $$r_\text{min-turn} \ = \ \frac{v^2}{g} \psi_\text{max},$$ where ${v}\in \mathbb{R}^2$ is the horizontal vehicle velocity, $\psi_\text{max}$ denotes the maximum bank angle and $g$ denotes the gravitational acceleration. It is desired to have $r_\text{l} > r_\text{min-turn}$ to be able to provide coverage while causing less physical strain on the UAV.

\noindent \textbf{Minimum loiter ($r_\text{l-min}$):} In the proposed approach, this is the loiter radius value chosen for each altitude. Ideally, $r_\text{l} > r_\text{min-turn}$ is better to reduce the physical wearing of the UAV in the long term, but the case of $r_\text{l-min} = r_\text{min-turn}$ results in the densest packing (minimum overlap) and the side length of the packing square is $a_\text{min} = r_\text{l-min}/ \sqrt{2}$. The radii at the higher flight levels are the multiples of this value.

\noindent \textbf{Coverage radius ($r_c$):} The coverage radius is the radius of the coverage footprint given the FOV and the instantaneous height($h$) of the vehicle. As seen from Fig.~\ref{fig:illustration}, it is defined as $$r_c \ = \ h \cdot \tan \theta.$$

\noindent \textbf{Communication radius ($r_\text{com}$):} Based on the on-board hardware, a UAV can connect to every other UAV within a certain distance, called the \textit{communication range}. Assuming an isotropic antenna for uniform range, the radius of the coverage is called the communication radius. Its value cannot be less than $\sqrt{2} r_\text{l-max}$, where $r_\text{l-max}$ is the loitering radius of the circle at the highest assigned level. It is a different entity from the loitering radius ($r_\text{l}$) and the sensor coverage radius ($r_\text{c}$). For any UAV $k$ at position $x_k$, its neighborhood is defined as $$\mathcal{N}_k \ \overset{\Delta}{=} \ \{x_i \in \mathbb{V}' \ | \ \text{dist}(x_k, x_i) \ \leq \ r_\text{com} \},$$ where $\text{dist}(x_k, x_i)$ represents the euclidean distance between the nodes $x_k$ and $x_i$.

\noindent \textbf{Sub- and super-square:} This represents the footprint squares for a particular altitude level. In Fig.~\ref{fig:illustration}, $a_1$ and $a_2$ are the side lengths of two different squares, for altitude levels $h_1$ and $h_2$ respectively. Here, $a_2$ represents the sub-square for $a_1$, and $a_1$ represents the super-square for $a_2$, where $a_2 \ = \ 2a_1$. When the UAV moves to a higher altitude level, it moves to the immediate super-square, and vice-versa. The process of moving the UAV from one altitude level to another is called \textit{level transition}. 

\noindent \textbf{Effective coverage ($E$):} It is defined as the total area covered by a loitering circle within the boundaries of the area of interest, minus the overlap with the immediate neighbors. These overlaps are purposefully introduced to allow the algorithm to cover every point (avoid coverage gaps) in the desired coverage area. The effective coverage for UAV k is defined as, $$E_k \ = \ (1-f) \pi r_l^2 \ - \ \sum_{i \in \mathcal{N}_k} A_{\text{ov-}i},$$ where $f$ is the fraction of the circle outside the area of interest and $A_{\text{ov-}i}$ is its overlap with the neighbor UAV $ i \in \mathcal{N}_{k}$.

\noindent \textbf{Full coverage:} It is defined as the state when each point in the area is guaranteed to be covered by at least one of the loitering UAVs at least once every loiter cycle during the operation. For $N$ UAVs deployed in the area $A$, it is achieved when $$A \  \subseteq \ \sum_{i=1}^{N} E_i. $$ This serves as the main objective of the presented work, where we adjust the radius of the loiter circle for the available UAVs to achieve full coverage.


\section{LEVEL-HOMOGENEOUS COVERAGE CONTROL}
\label{section:approach}
This section discusses the details of the proposed approach. The inputs from the end-user to the proposed approach is the coordinates of the vertices of the area to be covered by the UAV fleet, and the minimum loiter radius ($r_\text{l-min}$) of the vehicles. Based on these inputs, the algorithm starts by calculating the initial count of the nodes to be deployed and the locations on the initial (lowest) altitude level ($h_1$), which corresponds to the $r_\text{l-min}$, as defined. This process intrinsically marks the super-squares for different altitude levels as this is a top-down approach. After the initial deployment, the UAVs loiter around their respective circle at $h_1$. On failure of one or more agents in a sub-square, the failure handling mechanism is locally triggered and executed. Algorithm~\ref{alg1} summarizes the working of the proposed approach. The details of these processes have been discussed below.

\subsection{Initial Deployment}
\begin{figure}
    \centering
    \includegraphics[width=0.4\linewidth]{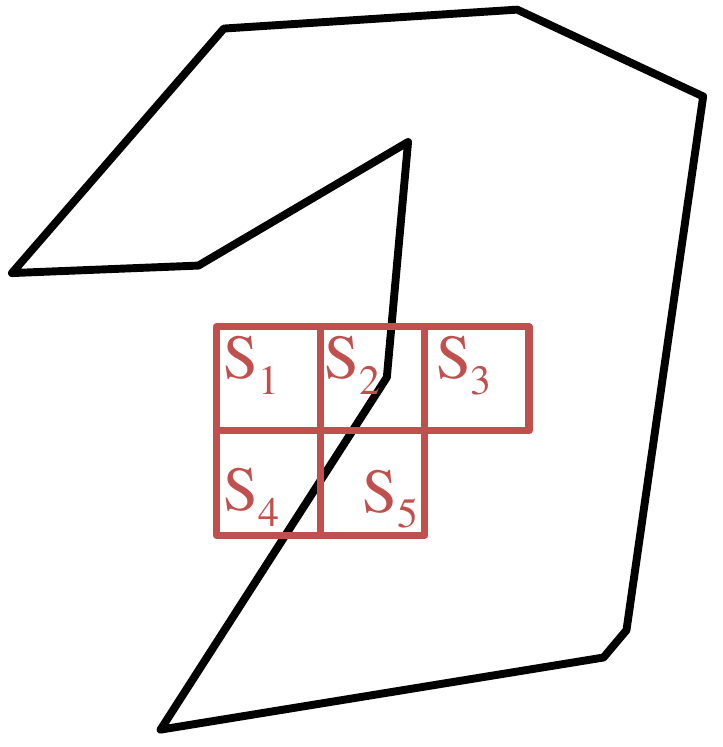}
    \caption{Illustration of how the algorithm classifies the squares during initial deployment. The black polygon marks the deployment boundary and the brown squares are the packed sample squares.}
    \label{fig:illustration_sq}
\end{figure}

This step employs a top-down approach for deciding upon the centres of the packing squares, but it is derived from $r_\text{l-min}$ to allow steps in altitude level and homogeneity in each of those levels. This results in $r_{\text{l}_1} \ = \ r_\text{l-min}$, $r_{\text{l}_2} \ = \ 2r_{\text{l}_1}$, and so on. This approach is limited to at most four altitude levels as the coverage quality at the $h_4$ level (based on $r_{\text{l}_4}$) is down to $(1/8)^{th}$ of that of the initially deployed $h_1$ level (based on $r_{\text{l}_1}$), which can be summarized as $$ r_{\text{l}_i} = 2^{i-1}r_\text{l-min} \text{, for } i \in \{1, \ 2, \ 3, \ 4 \}.$$ The deployment is carried out by bounding the area, by a large square ($r_\text{sq} \ = \ 16r_\text{l-min}$) and then bisecting the squares and narrow down to $r_\text{l-min}$. If $X_\text{area}$ and $Y_\text{area}$ are the arrays for $X-$ and $Y-$ coordinates of the area boundary vertices, a polygon surface $P_\text{area}$ is created as the region of interest, which is then used to check and classify if the squares at each altitude level lie inside or outside the desired coverage area with arbitrary geometry. We calculate the vertices of the aforementioned bounding square as follows
\begin{equation} 
\nonumber
\begin{split}
MIN_\text{sq} \ = \ & \min(\min(X_\text{area}), \min(Y_\text{area})), \\
MAX_\text{sq} \ = \ & 16 r_\text{l-min} + MIN_\text{sq}. 
\end{split}
\end{equation}
After the desired area is bounded by this square, it is bisected along both axes into four equal sub-squares. This process is continued till we reach the base level (level $h_1$). Following this, the list of squares on each level lying inside $P_\text{area}$ is created. The proposed algorithm uses a method that checks if any of the vertices of the squares in consideration lie inside $P_\text{area}$. If not, that square is discarded from the list. As for level $h_1$, this list gives the details and the count of the UAVs required to start with. This can be seen in Fig.~\ref{fig:illustration_sq}. Out of the five sample $h_1$ squares over the desired area, squares $S_2, \ S_3, \ S_4, \text{ and } S_5$ have 2, 4, 1 and 3 vertices inside $P_\text{area}$, respectively, and they are listed as UAV deployment locations. Square $S_1$ is completely outside $P_\text{area}$, and is discarded from the list. This is the last step for centralized planning after listing the deployment squares, where each UAV is given their center point to loiter at level $h_1$, and the UAVs fly to the respective loiter circles a by tracing the defined Dubin's path so as to keep them synchronised, by loitering in the same phase (shown in Fig.~\ref{fig:illustration}) at the same altitude level for any instance time instance. This synchronisation improves the instantaneous effective coverage by minimizing the overlap at that instant~\cite{sachin2020coordinated}. 

\subsection{Failure Detection}
Failure detection is a local phenomenon, when one or more agents suddenly drop out of operation. This could occur due to external impacts (e.g., blast), UAV instrument failure, power source failure or many other possible reasons, and the UAV is considered `non-recoverable' after the drop-out. As the UAVs are in communication with their super-square neighbors and with at least one other level-homogeneous sub-square UAV in neighboring super-square, at least two super-squares detect the failure. Based on the active communication links, each UAV maintains a list of the neighbors' state with all `1's. If a UAV drops out, its neighbors change the respective label to a zero, referring to `dropped-out'. That is, for a UAV $k$ with four neighbors, $N_k^\text{state} = [1, \ 1, \ 1, \ 1]$, for neighbors $N_k = [N_{k1}, \ N_{k2}, \ N_{k3}, \ N_{k4}]$ in all-active operation. If neighbor $N_{k3}$ drops out, the list is updated as $N_k^\text{state} = [1, \ 1, \ 0, \ 1]$. Furthermore, if $N_{k4}$ moves to a higher altitude level to restore coverage, it is updated to $N_k^\text{state} = [1, \ 1, \ 0, \ 2]$, denoting that it is loitering at the super-square radius on a higher altitude level, but is still online and connected. As the proposed failure handling approach tends to send one of the sub-square neighbors of the dropped-out UAV to a higher altitude level, the decision making is local and distributed within a super-square. If all four members of a super-square drop out, the neighboring super squares get involved in the decision making for restoring coverage. This will be discussed in detail below.

\begin{figure}[t!] 
\vspace{5mm}
    \centering
    \includegraphics[width=0.6\linewidth]{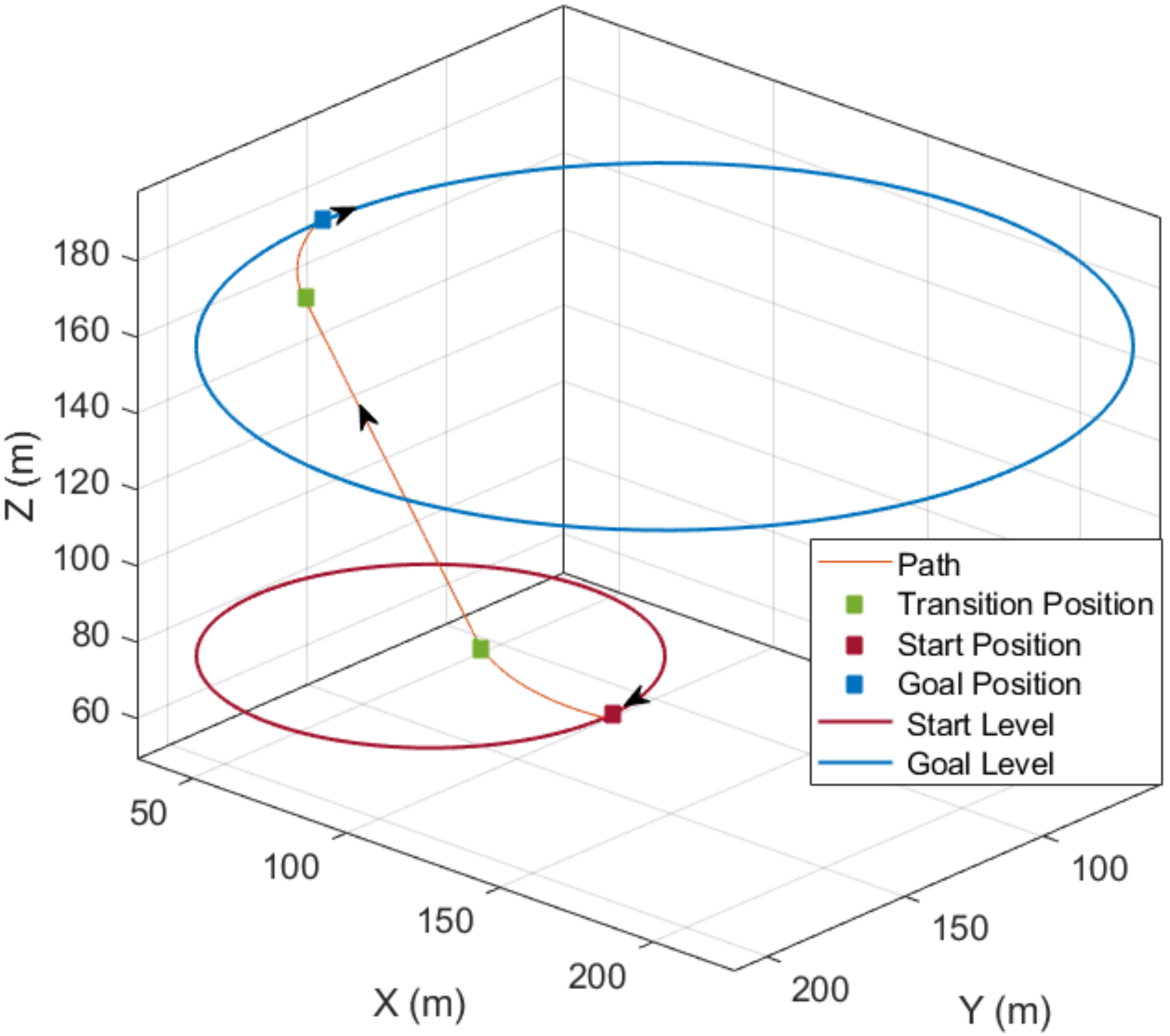}
    \caption{Transition of a UAV between two altitude levels through Dubin's path. The figure shows the transition path, start level, goal level, and the instantaneous headings are shown by arrows.}
    \label{fig:dubinexample}
\end{figure}

\begin{algorithm}[t!]
	\caption{Level-homogeneous deployment and recovery}

	\begin{algorithmic}[1]
		\Statex /* Initial Deployment */
		\Statex \textbf{Input:} $r_\text{l-min}, \ X_\text{area} \text { and } Y_\text{area}$
		\State Bounding square = \label{rep1}
		\Statex $MIN_\text{sq} \ = \ \min(\min(X_\text{area}), \min(Y_\text{area}))$
		\Statex $MAX_\text{sq} \ = \ 16 r_\text{l-min} + MIN_\text{sq}$
		\State Plot the boundary square and bisect it on both axes \label{rep2}
		\While {Side length, $ a > \sqrt{2} r_\text{l-min}$}
		    \State Repeat steps \ref{rep1} - \ref{rep2}
		\EndWhile
		\State Find Dubin's path to each loiter circle from base
		\State Deploy fixed-wing UAVs at $h_1$
		\Statex /* Failure Detection */
		\If {$dist(x,y) \ \leq \sqrt{2} r_\text{l-max}$}
		    \State UAV$_x$ and UAV$_y$ are connected
		    \If{ UAV$_x$ cannot connect to UAV$_y \ \forall x \in \mathcal{N}_{y}$}
		    \State UAV$_y$ dropped out
		    \EndIf
		\EndIf
		\Statex /* Failure handling (Recovery) */
		\If {$N_\text{new} \geq 4$} (Here, $h_4$, requiring 4 UAVs, is the highest allowed level before which quality gets compromised)
		    \For {each dropped-out node} \label{rep3}
		        \If{ No survivors in $N_{y \text{-sub-sq}}$}
		            \State Neighbor super-sq lends one for recovery
		            \State $x_1$ for recovering SuperSq of $y$ and $x_2$ to cover deficit of lending SuperSq
		            \State Calculate Dubin's path for $x_1$ and $x_2$ to respective loiter circles
		            \State Deploy UAV$_{x_1}$ and UAV$_{x_2}$ at $h_2$
		        \Else
		            \State Pick recovery UAV$_x \  \in  \ N_{y \text{-sub-sq}}$
		            \State Calculate Dubin's path for UAV$_x$ for SuperSq loiter circle
		            \State Deploy UAV$_{x}$ at $h_2$
	            \EndIf
	        \EndFor 
	        \State Full coverage restored \label{rep4}
	    \Else 
	        \State Full recovery not possible
	    \EndIf
		\State  Repeat Steps \ref{rep3}--\ref{rep4} for every level failure
	\end{algorithmic}
	\label{alg1}
\end{algorithm}

\subsection{Failure Handling}
When a failure and is detected, the algorithm aims to restore the coverage at the cost of the sensing quality of the coverage. It is intuitive that flying at a higher altitude level increases the coverage area, but decreases the quality. If the loiter radius is scaled along with the altitude level, it adds to the coverage time as well. Coverage time is the duration after which a point gets covered by a loitering UAV in every loiter cycle.

The proposed algorithm can solve the cases of single, simultaneous multiple, clustered and spread out nodes failures efficiently as it uses a distributed approach where the failure detection and recovery decision making is limited to the minimum possible number of neighbors in the failure area. For failure of less than three sub-square loitering UAVs, the decision making is limited within that super-square to minimize recovery time. The nearest sub-square survivor UAV (in terms of phase, to minimize the transfer path length and time) breaks from its loiter circle to move to the super-square loitering at a higher altitude level, as explained in the following paragraph. If only one of the four sub-square UAVs has survived, it moves to a higher altitude level by default. In a more complicated loss scenario, there could be no sub-square survivors in a super-square, for all four having dropped out, or the super-square being at the edge of the boundary and thus no backups for coverage restoration. In this case, the neighboring super-squares will fill in as they have also detected the occurrence and the extent of the failure. Aiming for the shortest transfer path based on phase and heading, the two chosen sub-square UAVs from a neighbouring super-squares break from their loiter circles and trace Dubin's path to the higher altitude loitering circles, one each for their own and the failed neighboring super-square. It is worth mentioning again that the loitering motions after moving to the higher altitude level have to be synchronised in phase as well, as the proposed approach solves for `level-homogeneous' and it improves the effective coverage at that level.

\subsection{Dubin's Path}
The recovery UAV has to break off from its current loiter circle at some point to trace a Dubin's path to the loiter circle of a larger radius at a higher altitude level and join in at a prescribed point, to maintain synchronization. This can be achieved by controlling the break-off point, the join-in point, and the headings at both points, and the motion of the UAV. Typically, Dubin's paths are created as combinations of circular sections and straight lines, with an aim to minimize the travel time and distance. In 3D cases like this, these can form helical structures while moving to a higher or lower altitude level based on the height difference. The motion of the UAV is constrained into six options: straight (S), left turn (L), right turn (R), helix left turn (Hl), helix right turn (Hr), and no motion (N). The equations of motion and the generation of Dubin's paths have been followed from ~\cite{savla2007coverage, mclain2014implementing, lugo2014dubins}. There could be cases when some of these motion combinations might be physically restricted for the UAVs (e.g. LSLN). An example for Dubin's path transition of a UAV is shown in Fig.~\ref{fig:dubinexample}. The starting and final (goal) altitude levels and loitering circles have been shown, along with the path with a minimum turning radius of 80 metres. The instantaneous headings and positions of the UAV have been marked by arrows. It can be seen that the path is defined by RSRN motion and the start (break off), the goal (join-in) and the transition (motion change) positions have been marked. For multiple UAVs, the path lengths and the join-in point are to be controlled to achieve and maintain synchronization.

\section{SIMULATION ANALYSIS}
\label{section:sims}
\begin{figure*}
	\centering
	\begin{subfigure}[b]{0.48\textwidth}
	\includegraphics[width=1\linewidth]{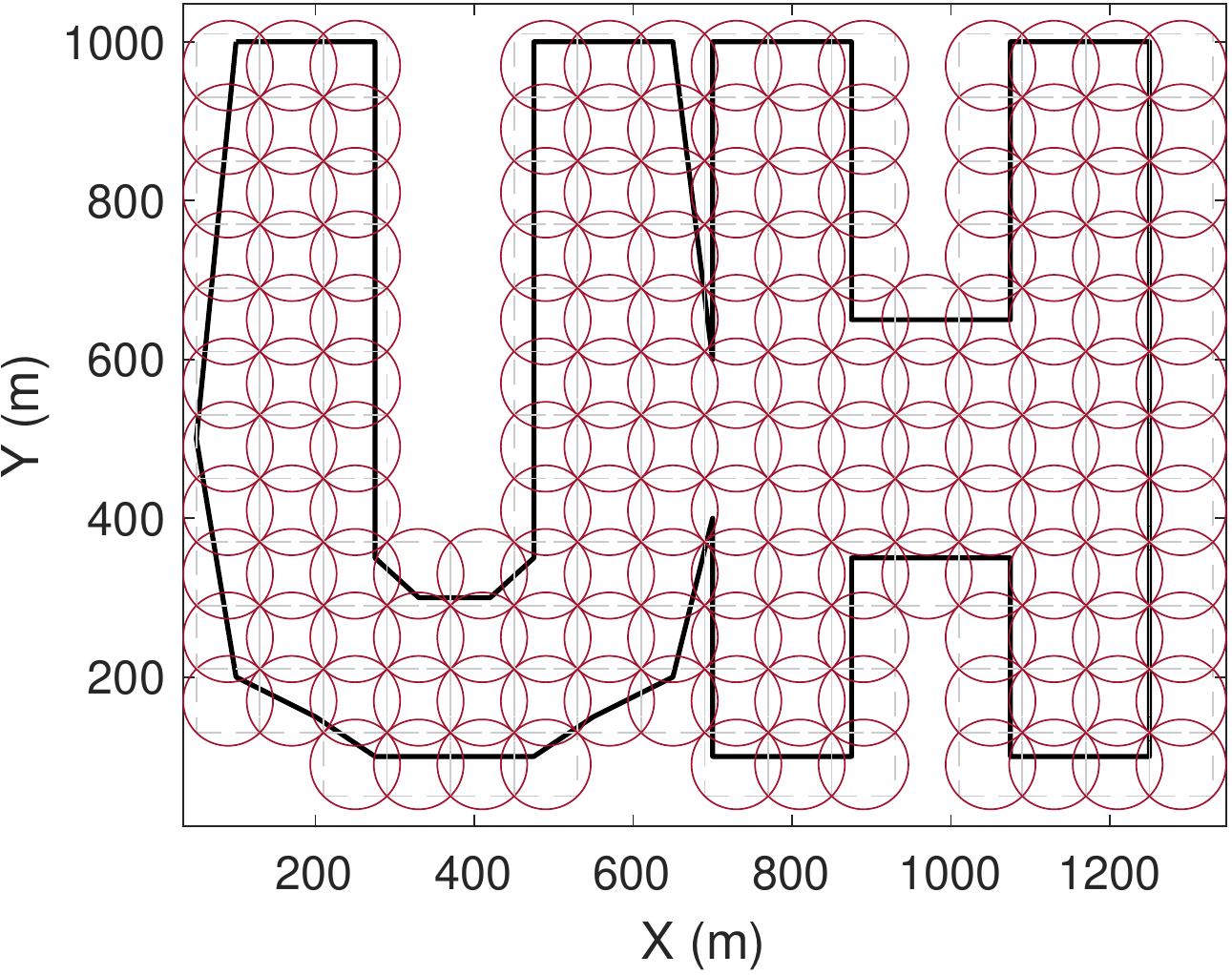}
	\caption*{(a)}
	\end{subfigure}
		\vspace{3mm}
	\begin{subfigure}[b]{.48\textwidth}
    \includegraphics[width=1\linewidth]{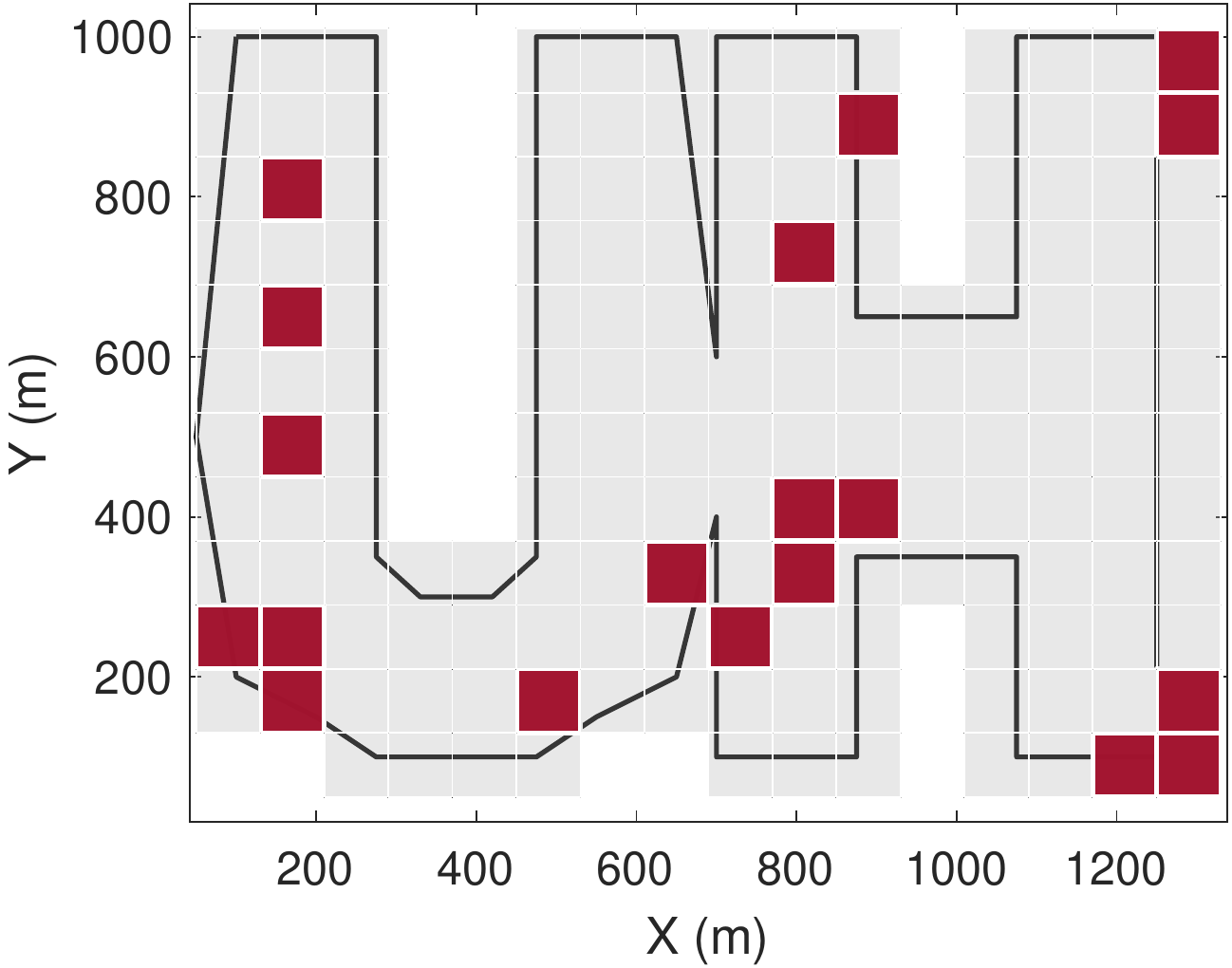}	
    \caption*{(b)}
	\end{subfigure}
	\begin{subfigure}[b]{.48\textwidth}
    \includegraphics[width=1\linewidth]{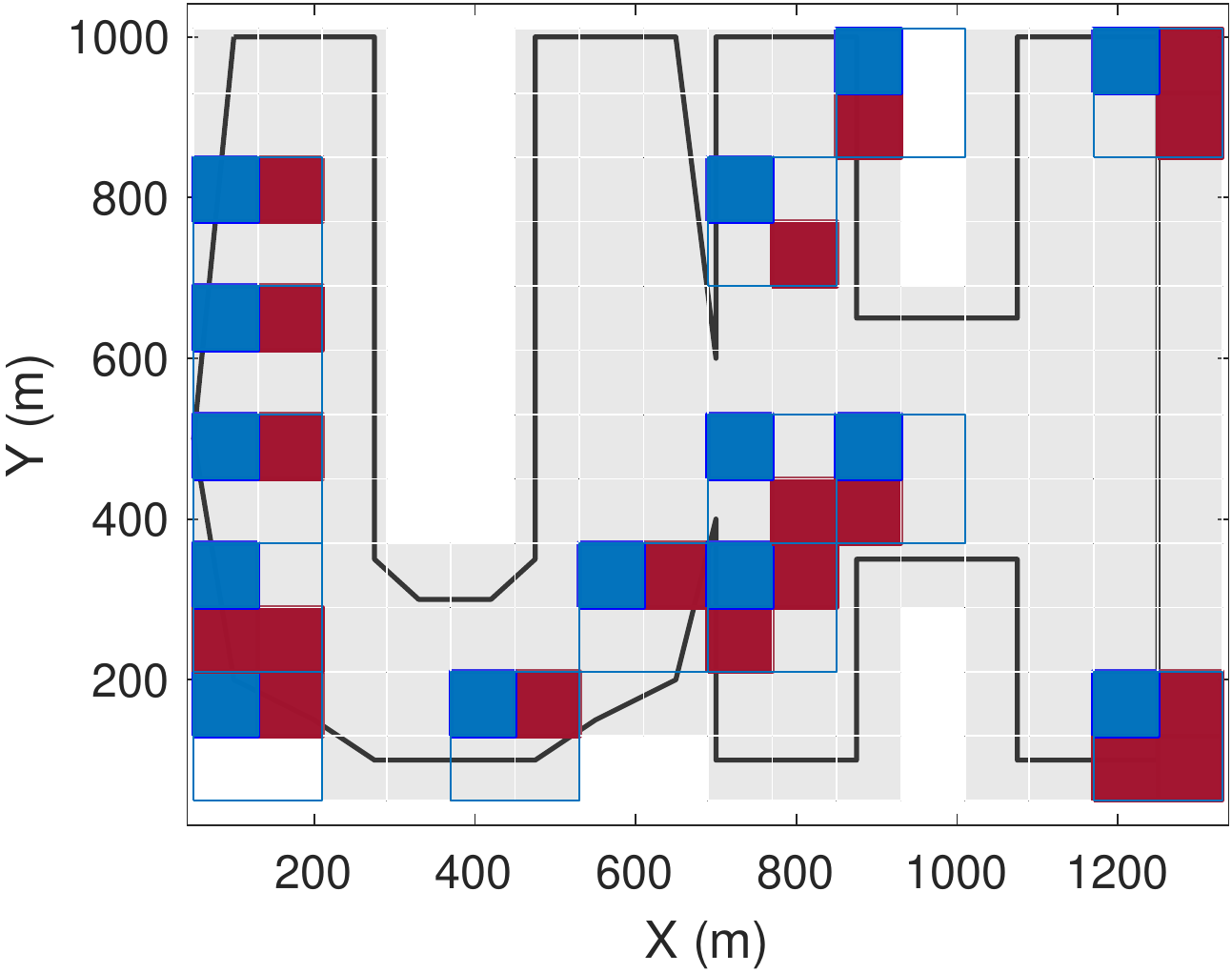}	
    \caption*{(c)}
	\end{subfigure}
	\begin{subfigure}[b]{.48\textwidth}
    \includegraphics[width=1\linewidth]{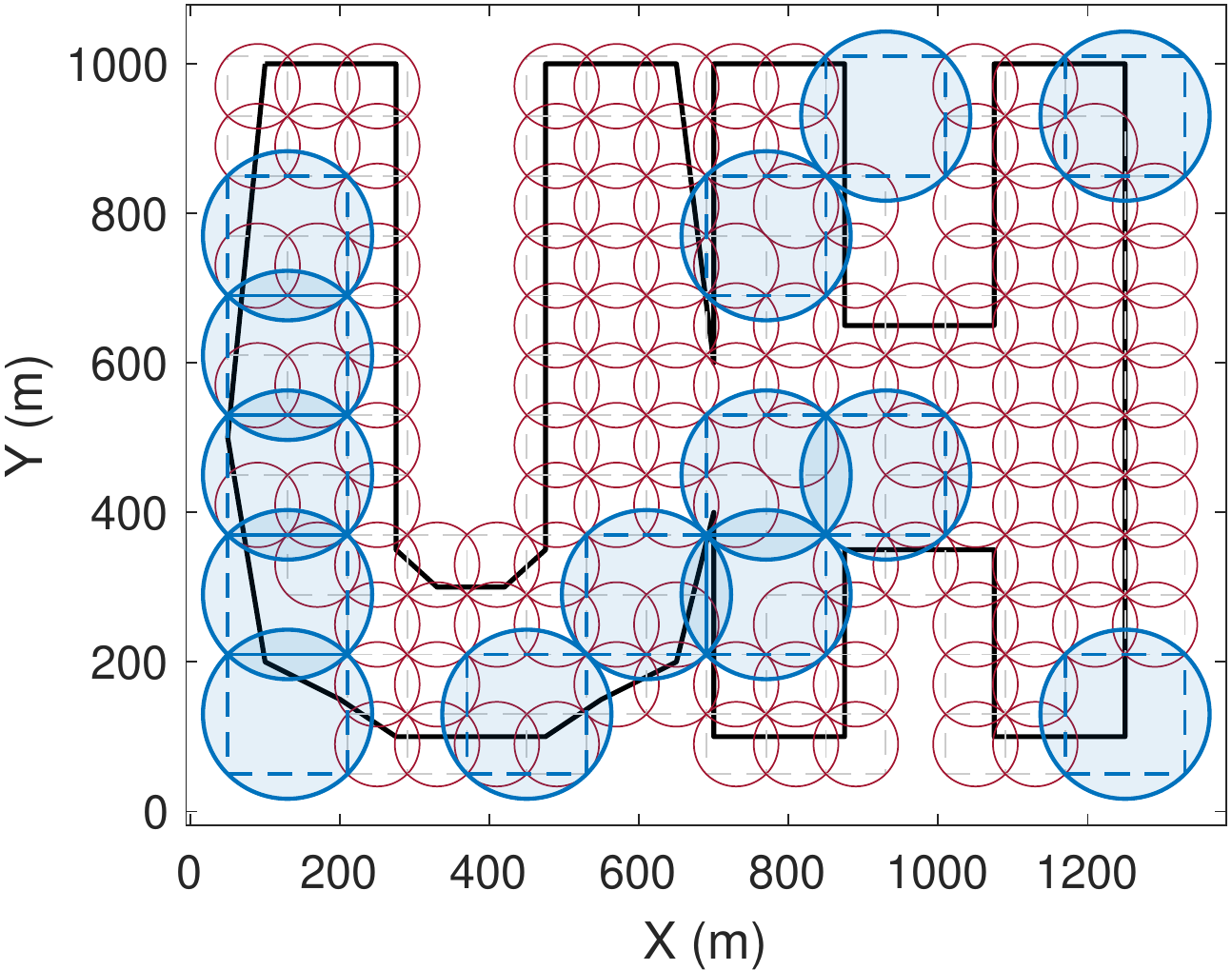}	
    \caption*{(d)}
	\end{subfigure}
	\caption{Simulation of the proposed approach for resilient coverage after agents' drop out to maintain full coverage: (a) Initial deployed network with minimum loiter radii; (b) Simultaneous multiple node loss spread over the area; (c) Full coverage recovery by deploying sub-square neighbors (blue) of the dropped-out nodes (brown) at a higher altitude level, with larger loiter radii; (d) Final network layout with different sized loiter circles at different altitude levels.}
	\label{fig:simulation}
\end{figure*}

The proposed algorithm was applied to various arbitrary shaped areas, while varying the minimum loiter radius ($r_\text{l-min}$), to verify it applicability. Simulation results for one such area and $r_\text{l-min}$ setup have been shown in Fig.~\ref{fig:simulation}. To show that the algorithm works perfectly for difficult non-convex area shapes (and to be a bit geeky), the area shape has been made to look like UH, which stands for University of Hawai'i in this case. The parameters used for the simulation have been listed in Table.~\ref{table:simdata}. The arbitrary geometric area boundary has been shown as the black polygon. The value of FOV of the sensor has been taken to be 45$\si{\degree}$ and this results in the coverage radius equal to the altitude level and $$\frac{r_\text{c1}}{r_\text{c2}} = \frac{h_{1}}{h_{2}}.$$

Fig.~\ref{fig:simulation}(a) shows the initial deployed UAV network at altitude level $h_1$ with loiter radius $r_\text{l-min}$. The packing squares (shown in dashed grey lines) are inscribed by the respective loitering circles (shown in red), where each circle represents the path for a UAV. It can be seen that the area is fully covered using the least number of loitering UAVs while skipping the undesired areas. 

\begin{table}
	\caption{Parameters used in the simulation case shown in Fig.~\ref{fig:simulation}.}
	\centering
	\begin{tabular}{|l|l|}
		\hline 
		$X_\text{area}$ (m) & [100, 50, 100, 200, 275, 475, 550, 650, 700, 700, 875, 875, 1075, 1075, 1250, 1250, \\
		 &  1075, 1075, 875, 875, 700, 700, 650, 475, 475, 420, 330, 275, 275, 100] \\ 
		\hline 
		$Y_\text{area}$ (m) & [1000, 500, 200, 150, 100, 100, 150, 200, 400, 100, 100, 350, 350, 100, 100, 1000, \\
		 &  1000, 650, 650, 1000, 1000, 600, 1000, 1000, 350, 300, 300, 350, 1000, 1000]  \\
		\hline 
		Loiter radius (m) & $80$ \\ 
		\hline 
		Bounding Square (m$^2$) & 1330 $\times$ 980 \\ 
		\hline 
		Number of Initial Nodes & 109 \\
		\hline 
		Number of Lost Nodes & 19 \\
		\hline 
		FOV & 45$\si{\degree}$ \\
		\hline 
	\end{tabular}
	\label{table:simdata}
\end{table}

In Fig.~\ref{fig:simulation}(b), a random node loss scenario is applied, which results in node dropouts spreading across the area (marked by solid brown squares). Ten of the super-squares have lost one sub-square node each, three of them have lost two each and one of them has lost three sub-square nodes.  The gray shaded squares represent still active UAVs after the dropout, which are loitering on circles inscribing those squares. The circles have not been shown to make the plot less crowded. These failed nodes, each being considered a sub-square, belong to various super-squares across the area. At this point, the algorithm aims to resolve the issue locally and restore the coverage by sending some of the sub-square nodes from the super-squares which have lost members to altitude level $h_2$. 

\begin{figure}[t!]
\vspace{5mm}
    \centering
    \includegraphics[width=0.7\linewidth]{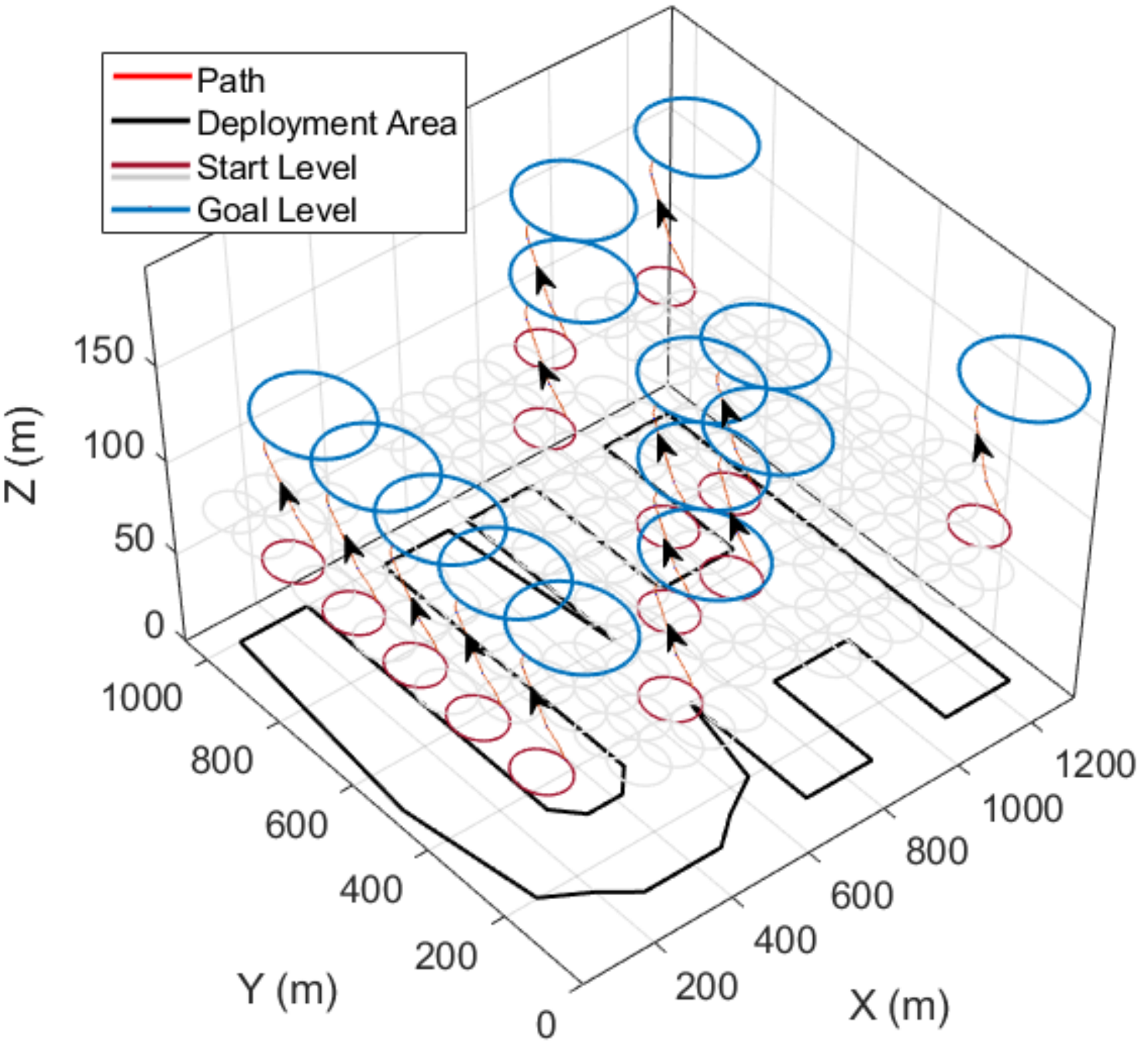}
    \caption{Transition of multiple UAVs between two levels through Dubin's path to recover the full coverage from a simultaneous multiple node loss scenario shown in Fig.~\ref{fig:simulation}. The headings of the UAVs during the transit have been marked by arrows.}
    \label{fig:dubin}
\end{figure}

After the algorithm is applied, the resolved network has been shown in Fig.~\ref{fig:simulation}(c). The lost nodes have been shown in solid light brown and the chosen recovery UAVs have been shown in solid blue. The new trajectories for the recovery UAVs are the loiter circles inscribing the blue squares shown in the figure. It can be seen that each super-square which had lost one or more of their sub-square nodes chose one survivor sub-square node and sent it to the altitude level $h_2$ to loiter at radius $r_{\text{l}_2} \ = \ 2r_\text{l-min}$. The new loiter circles for the active nodes have been shown in Fig.~\ref{fig:simulation}(d). 

Fig.~\ref{fig:dubin} shows how all the assigned UAVs move from a lower altitude level ($h_1$= 80 metres in this case) to a higher one ($h_2$= 160 metres) using Dubin's paths. The desired coverage area has been shown in black. The start positions are on the lower loiter circles (shown in brown) of radius  $r_{\text{l}_1} \ = \ r_\text{l-min}$ and the point where they join are on the larger loiter circles (shown in blue) $r_{\text{l}_2} \ = \ 2r_\text{l-min}$. The gray circles at the lower level $h_1$ do not go under the transition and continue loitering at the same level. The effect of the turn radius constraint can be seen in the plot. In cases where these circles are far apart in height, the UAVs trace helical paths to get to the target loitering circle at a higher or a lower altitude. All the assigned sub-square loiter UAVs move to altitude level $h_2$ for synchronised loitering and coverage, and this completes the recovery process. 

\section{CONCLUSION}
\label{section:last}
This paper presents an approach to deploy and maintain a network of fixed-wing unmanned aerial vehicles for persistent coverage over an area. This is to exploit the superiority of fixed-wing vehicles over traditionally used rotor-type vehicles in terms of endurance, making them better suited for long-term and large scale deployment problems. The initial locations for deployment over an arbitrary geometric area are generated by adaptive square packing, and failures of nodes are handled by flying at different altitude levels as required. This is a distributed approach for failure detection and handling, and is scalable for various area sizes. It can also adapt to difficult deployment area geometry and vehicles with various specifications. Simulation results show that the algorithm is applicable to all sorts of arbitrary geometry of an area, and can also be implemented by various UAVs with diverse physical properties and constraints.

A future research direction for this work is the experimental verification of the proposed algorithm, using standard mini-UAV platforms, by varying fleet size and minimum loiter radius. Another research direction is to modify the algorithm to implement it with heterogeneous UAVs flying at the same altitude level, and to introduce the weights based on coverage information importance within the deployment area. 

\bibliography{ran}

\end{document}